\documentclass{article}

 \usepackage[preprint]{neurips_2026}

\usepackage[utf8]{inputenc} 
\usepackage[T1]{fontenc}    
\usepackage{hyperref}       
\usepackage{url}            
\usepackage{booktabs}       
\usepackage{amsfonts}       
\usepackage{nicefrac}       
\usepackage{microtype}      
\usepackage{xcolor}         
\usepackage{graphicx} 
\usepackage{enumitem}
\usepackage{amsmath}
\usepackage{wrapfig}
\usepackage{booktabs}
\usepackage{multirow}
\usepackage{tabularx}
\usepackage{array}
\usepackage{makecell}
\usepackage[most]{tcolorbox}
\usepackage{fvextra}
\usepackage{algorithm}
\usepackage{algorithmic}
\title{Mechanistic Attention Guidance for \\ Agent Memory Refinement}

\author{%
  Yechao Hong\textsuperscript{1}\thanks{Equal contribution.},
  Haiquan Qiu\textsuperscript{1}\footnotemark[1],
  Yaqing Wang\textsuperscript{2},
  Quanming Yao\textsuperscript{1}\thanks{Corresponding author.} \\
  \textsuperscript{1}Department of Electronic Engineering, Tsinghua University \\
  \textsuperscript{2}Beijing Institute of Mathematical Sciences and Applications \\
  \texttt{hongyechao57@gmail.com}
}

\begin{document}

\maketitle

\begin{abstract}
Existing self-evolving memory systems mainly improve agent memory based on textual outputs, such as task trajectories and reflections.
However, this text-based paradigm rarely incorporates internal mechanistic signals, leaving how retrieved memory is actually utilized during task execution underexplored.
This limitation can lead to unreliable error attribution and hallucinated memory modifications.
In this work, we show that retrieval-head attention provides a mechanistic signal for revealing segment-level memory utilization.
By aggregating attention over memory segments and decision steps, we construct a context utilization matrix that exposes recurring memory-use patterns and indicates corresponding refinement strategies.
Building on this observation, we propose \textbf{Attention-Guided Memory Refinement (AGMR)}, a framework that uses utilization patterns revealed by attention to guide targeted segment-level memory updates.
AGMR corrects or enhances memory for failed executions, simplifies memory for successful executions, and verifies each update through re-execution.
Experiments on interactive decision-making benchmarks show that AGMR improves both task performance and memory efficiency over text-only memory refinement baselines.
Code is available at \url{https://anonymous.4open.science/r/AGMR_code-3262/}.
\end{abstract}

\section{Introduction}

\begin{figure}[hbt]  
  \centering
  \vspace{-10pt}
  \includegraphics[width=\linewidth]{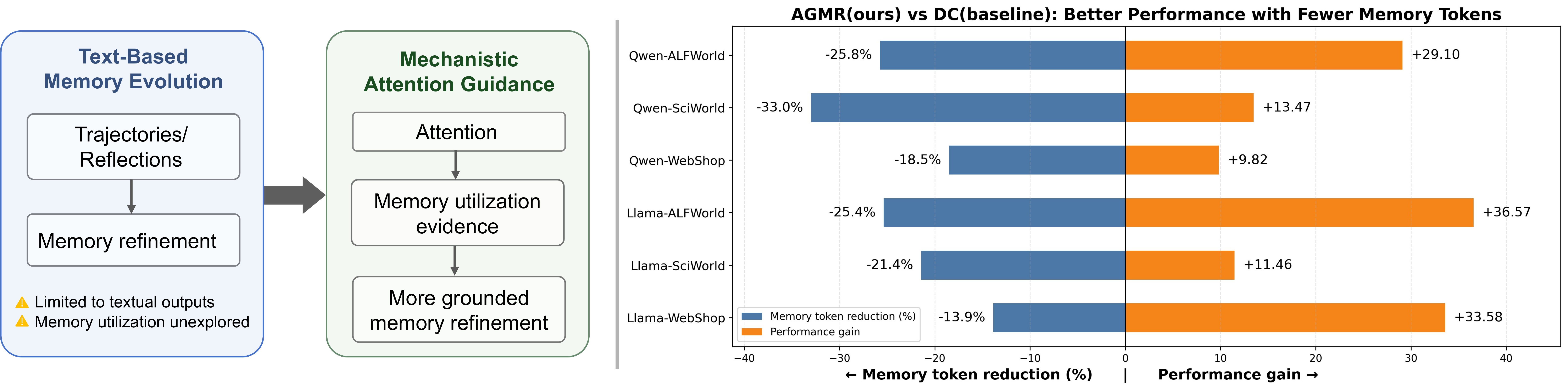}\caption{
Mechanistic attention guidance enables more grounded memory refinement, leading to higher performance and fewer memory tokens compared to baseline.
}
  \vspace*{-8pt}  
  \label{fig:framework}
\end{figure}

Large language model (LLM) agents have shown strong potential in interactive and long-horizon decision-making by coupling reasoning, action generation, and external memory~\cite{yao2022react, park2023generative, wang2023voyager}. 
Memory allows agents to accumulate experience from past interactions and reuse it in future tasks. 
Prior work has explored various memory forms, including trajectories, workflow abstractions, procedural knowledge, and long-term personal experience~\cite{zheng2023synapse, wang2024awm, zhong2024memorybank, chhikara2025mem0, fang2025memp, wang2025rememberme, ouyang2025reasoningbank}. 
More recently, memory has been treated as a self-evolving component that is continuously updated through task experience~\cite{shinn2023reflexion, zhao2024expel, suzgun2026dynamic, zhang2025ace, cai2025flex, fang2026trajectory, yan2025memoryr1}.

Despite this progress, existing self-evolving memory systems remain largely driven by \emph{model-generated text}, such as task trajectories and reflective analyses.
They typically execute a task, record a textual trajectory, and then use model-generated text to attribute errors, extract experience, and rewrite memory~\cite{shinn2023reflexion, zhao2024expel, suzgun2026dynamic, zhang2025ace, cai2025flex}.
While effective, this paradigm leaves two important aspects underexplored.
First, it relies primarily on textual outputs while making limited use of internal mechanistic signals.
Task trajectories record what the model says and does, but not what drives its decisions.
This concern is consistent with findings that LLM-generated explanations can be plausible yet unfaithful, often acting as post-hoc rationalizations rather than faithful accounts of model behavior~\cite{turpin2023language, agarwal2024faithfulness, chuang2026faithlm, arcuschin2025cotwild}.
Consequently, model-generated text can be a weak optimization signal for memory evolution and an unreliable basis for error attribution.
Second, prior work mainly studies how memory should be organized and updated~\cite{wang2024awm, chhikara2025mem0, fang2025memp, ouyang2025reasoningbank, wang2025rememberme}, but rarely asks how retrieved memory is actually utilized during task execution.
Thus, memory refinement remains largely a free-form reflection process over text, performed without a structured evidence framework or external anchor for deciding which memory segments should be updated and how.
Together, these factors can lead to unreliable error attribution and hallucinated memory modifications~\cite{sharma2023sycophancy, kalai2025whyllmshallucinate, song2025hallucinationtax}.

To address these limitations, we leverage retrieval-head attention as a mechanistic signal for memory utilization.
Recent mechanistic interpretability studies suggest that attention heads can exhibit functional specialization, with some heads playing important roles in contextual retrieval and information routing~\cite{wu2025retrievalheads, kahardipraja2025atlasicl, bick2025gather, zhang2025query}.
Motivated by these findings, we aggregate attention from retrieval-related heads over memory segments and decision steps to construct a \textbf{context utilization matrix}, which provides structured evidence of how memory is actually used during task execution.
This matrix reveals recurring memory-use patterns: for failed tasks, it distinguishes whether the agent attends to misleading memory, misses relevant memory, or is distracted by irrelevant memory; for successful tasks, it exposes weakly utilized redundant memory that can be simplified.
These patterns turn retrieval-head attention into mechanistic guidance for agent memory refinement, linking memory updates to internal evidence beyond text-only reflection.

Building on this observation, we propose \textbf{Attention-Guided Memory Refinement (AGMR)}, a framework that uses attention-revealed utilization patterns to guide targeted segment-level memory updates.
For failed executions, AGMR corrects misleading memory, enhances missed relevant memory, or addresses distracting memory.
For successful executions, it simplifies semantically redundant and weakly utilized memory content.
Each candidate update is further validated through re-execution before being committed to the memory base.
Experiments on interactive decision-making benchmarks show that AGMR improves both task performance and memory efficiency over text-only memory refinement baselines.

\section{Attention as a mechanistic signal for memory utilization}
\label{sec:key_insight}

\subsection{Extracting utilization signals from attention}
\label{sec:context_utilization_trajectory}

\paragraph{Retrieval head selection.}
Textual trajectories describe the agent's outputs but do not show how memory is accessed during generation.  
Motivated by prior findings that some attention heads are involved in contextual retrieval and information routing, we use attention from these retrieval-related heads as a proxy for segment-level memory utilization.  
Since not all heads reflect memory access, we first identify retrieval heads that selectively attend to task-relevant contextual information. 
These heads are then used to track attention over memory segments during execution.

We construct a retrieval supervision set 
$\mathcal{D}_{\mathrm{ret}}=\{(\texttt{key\_context}^{(i)},\texttt{formatted\_prompt}^{(i)})\}_{i=1}^{N}$ for each model, 
where \texttt{formatted\_prompt}$^{(i)}$ is the complete model input immediately before generating a key decision step in a task, 
and \texttt{key\_context}$^{(i)}$ is the memory snippet that should ideally be retrieved to support that step. 
Dataset details are provided in Appendix~\ref{app:retrieval_dataset}. 
To identify retrieval heads, we follow the approach of \cite{zhang2025query}, using attention from query tokens as a proxy for relevant-memory retrieval; in our interactive agentic tasks, the query naturally corresponds to the last token of \texttt{formatted\_prompt}$^{(i)}$.
We therefore score each head by the total attention from the last token of \texttt{formatted\_prompt}$^{(i)}$ to the corresponding \texttt{key\_context}$^{(i)}$ tokens, and then aggregate this score over all supervision pairs:
\[
\mathrm{score}_{l,h}
=
\sum\nolimits_{i=1}^{N}
\sum\nolimits_{t \in \texttt{key\_context}^{(i)}}
\mathrm{Attention}_{l,h}^{(i)}(\mathrm{last\_token}^{(i)} \rightarrow t).
\]
Higher-scoring heads are more likely to retrieve task-relevant memory. 
For each backbone model, we select the top 5 scoring heads as the retrieval-head set $\mathcal{H}_{\mathrm{ret}}$, which is used to construct context utilization matrix.
The full score distributions are provided in Figure~\ref{fig:retrieval_score_heatmap} and Appendix~\ref{app:retrieval_score}.

\begin{figure}[hbt]  
  \centering
  \includegraphics[width=\linewidth]{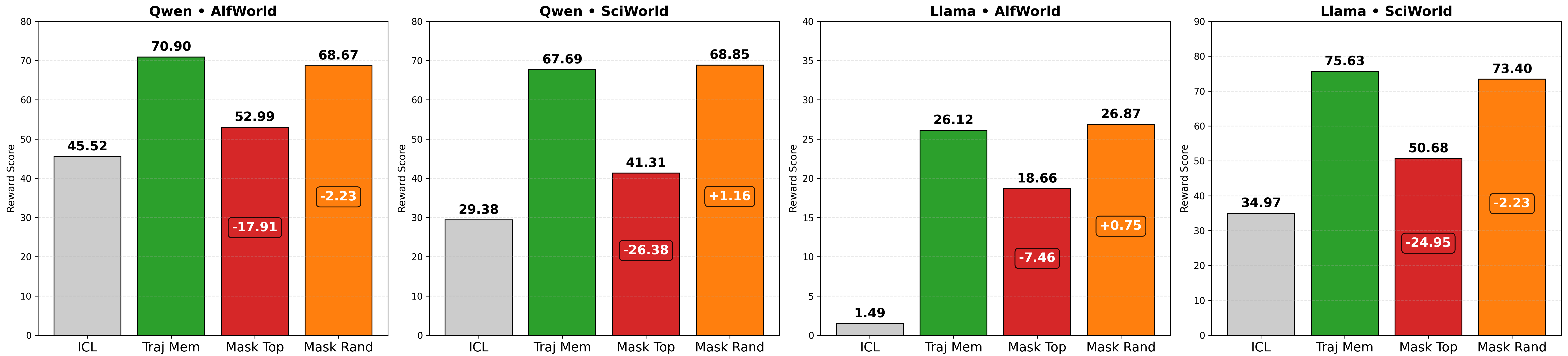}\vspace{-6pt}\caption{
Ablation validation of identified retrieval heads. 
\texttt{ICL} is the no-memory baseline, while \texttt{Traj Mem} adds retrieved trajectory memory and yields substantial gains. 
Masking high-score heads (\texttt{Mask Top}) largely reduces the memory-induced gains, whereas masking random heads (\texttt{Mask Rand}) has minor effect. 
Numbers inside the bars indicate reward changes relative to \texttt{Traj Mem}.
}
  \vspace{-6pt}
  \label{fig:retrieval_head_ablation}
\end{figure}

We validate the selected retrieval heads by masking them under the raw trajectory-memory setting on ALFWorld and ScienceWorld (abbreviated as SciWorld hereafter). 
Specifically, we mask the top 10 scoring heads for Qwen-2.5-7B-Instruct and the top 5 scoring heads for Llama-3.1-8B-Instruct, with randomly selected heads as controls. 
As shown in Figure~\ref{fig:retrieval_head_ablation}, adding trajectory memory substantially improves performance, but masking the top-scoring heads largely removes this gain, whereas masking random heads has only a minor effect.
This supports their functional role in retrieving useful memory information.


\paragraph{Context utilization matrix construction.}
To capture how the agent actually utilizes memory during execution, we aggregate attention from the selected retrieval heads over memory segments at each decision step. Given a retrieved trajectory memory \(m^*\), we segment it into an ordered list \(m^*=[s_0,\dots,s_L]\), where each segment \(s_\ell\) contains the observation, thought, and action from one consecutive trajectory step.
The agent executes the task with \(m^*\) and produces a step list \(\tau=[u_0,\dots,u_T]\).

\begin{wrapfigure}[18]{r}{0.42\linewidth}  
  \centering
  \vspace{-10pt}
  \includegraphics[width=\linewidth]{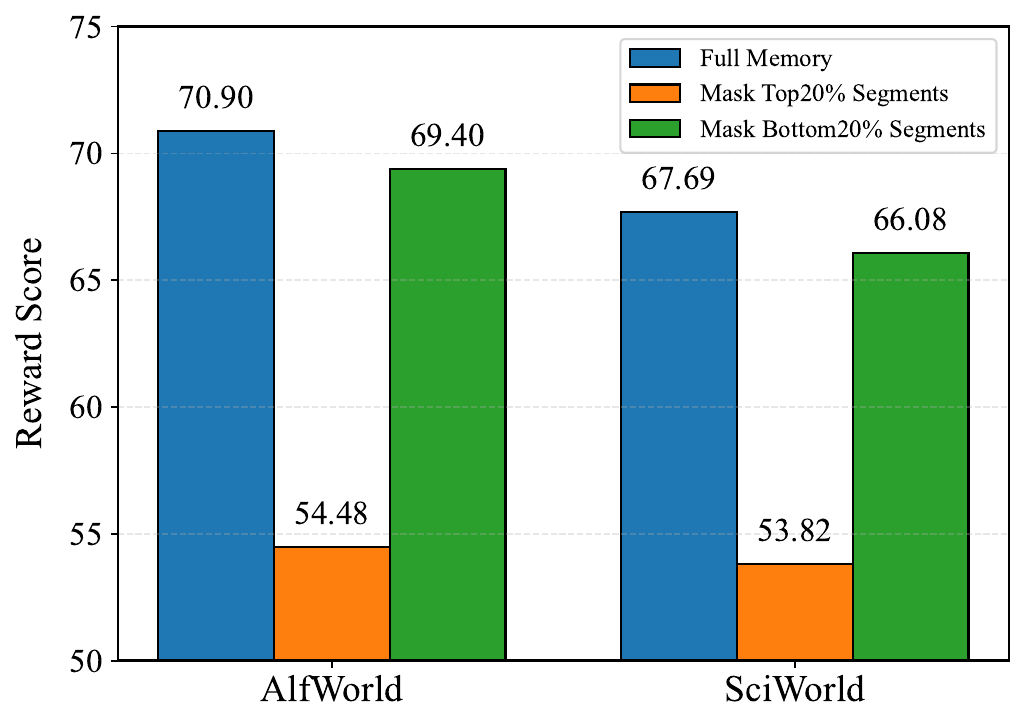}\vspace{-6pt}\caption{
Masking high-attention memory segments causes a much larger performance drop than masking low-attention segments, supporting attention strength as a proxy for memory utilization.
}
  \label{fig:masking_segments}
\end{wrapfigure}

Because observations can be extremely long and vary substantially across segments, we define $\bar{s}_\ell \subseteq s_\ell$ by excluding observation tokens from segment $s_\ell$.
For the generation of the $t$-th step $u_t$, let $q_t$ be the last prompt token immediately before generating $u_t$.
For each retrieval head $h \in \mathcal{H}_{\mathrm{ret}}$ and segment $s_\ell$, we aggregate attention from $q_t$ to the tokens in $\bar{s}_\ell$.
Since the prompt grows during execution and attention is normalized over a longer context, we rescale the score by the current prompt length $n_t$.
The final step-level utilization score for segment $s_\ell$ is
\[
\tilde{a}_{t,\ell}
=
\frac{n_t}{|\mathcal{H}_{\mathrm{ret}}|}
\sum\nolimits_{h\in\mathcal{H}_{\mathrm{ret}}}
\sum\nolimits_{j\in\bar{s}_\ell}
\mathrm{Attention}_{h}(q_t\rightarrow j).
\]
Computing this score for all memory segments yields a step-level utilization vector
\(\tilde{\mathbf{a}}_t=[\tilde{a}_{t,0},\dots,\tilde{a}_{t,L}]^\top\).
Repeating this process for all decision steps yields the \textbf{context utilization matrix}
\[
\mathbf{C}
=
[\tilde{\mathbf{a}}_0,\tilde{\mathbf{a}}_1,\dots,\tilde{\mathbf{a}}_T]
\in\mathbb{R}^{(L+1)\times(T+1)},
\]
where each entry $\mathbf{C}_{\ell,t}$ is an attention-derived estimate of how strongly segment $s_\ell$ is utilized before generating step $u_t$.

To examine whether this attention-derived matrix provides a useful proxy for memory utilization, we conduct a masking test on Qwen-2.5-7B-Instruct.
For each retrieved segmented memory entry, we rank segments by their maximum utilization across all execution steps,
\(c^{\max}_\ell=\max_t \mathbf{C}_{\ell,t}\), the maximum value in row \(\ell\) of \(\mathbf{C}\), and remove either the top 20\% or bottom 20\% segments.
As shown in Figure~\ref{fig:masking_segments}, removing high-$c^{\max}_\ell$ segments causes a clear performance drop, while removing low-$c^{\max}_\ell$ segments has little effect.
This contrast suggests that high-scoring memory segments are more useful for task execution, supporting $\mathbf{C}$ as a mechanistic signal for memory utilization.

\subsection{Different memory-utilization patterns revealed by attention}
\label{sec:different_patterns}
\begin{figure}[t]  
  \centering
  \includegraphics[width=\linewidth]{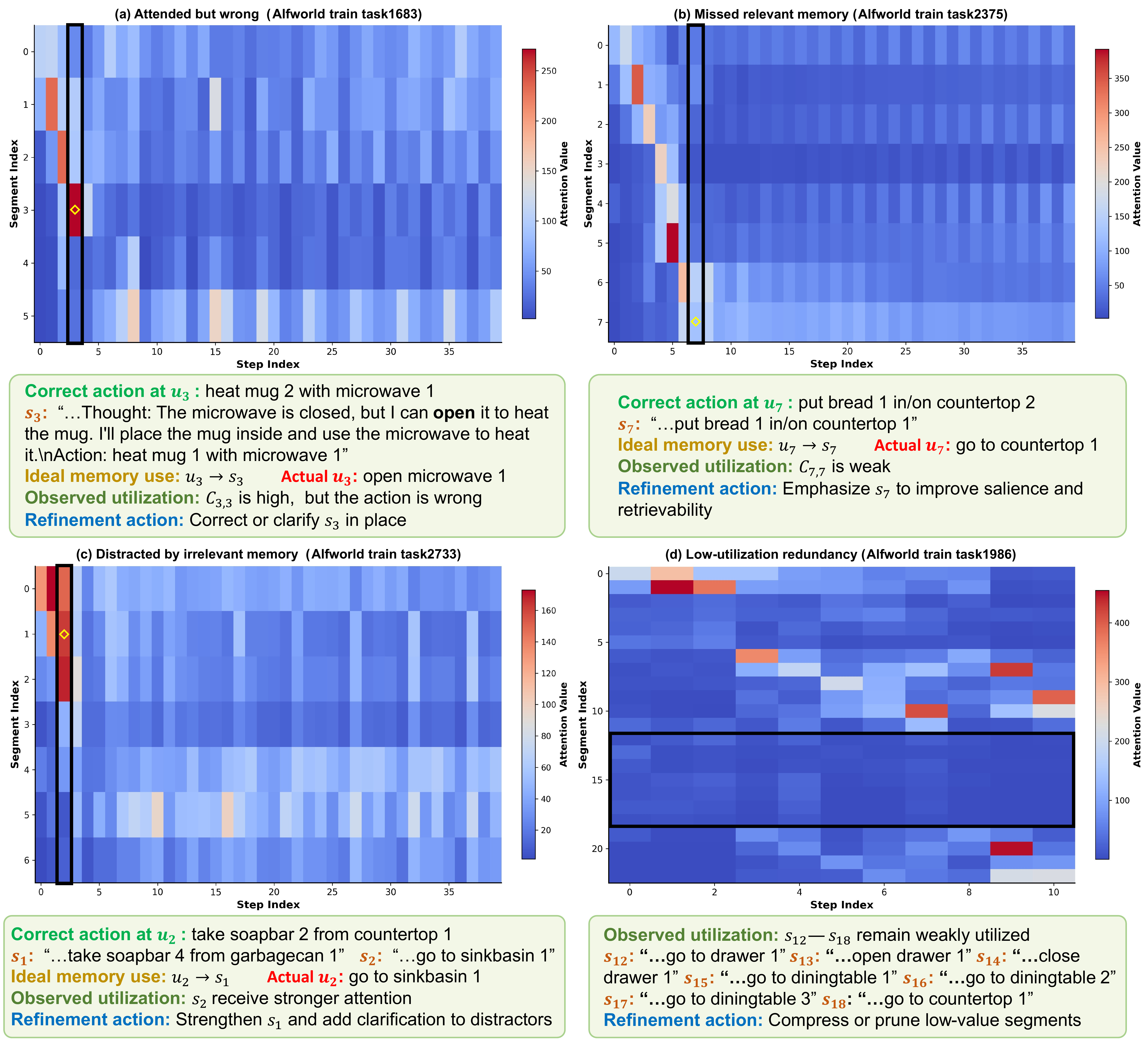}\vspace{-6pt}\caption{
Context utilization matrices of representative cases: (a) attended but wrong, (b) missed relevant memory, (c) distracted by irrelevant memory, and (d) low-utilization redundancy.
Here, \(s_\ell\) denotes the content of the \(\ell\)-th memory segment, and \(u_t\) denotes the agent output at the \(t\)-th step; in failure cases, ``actual \(u_t\)'' denotes the erroneous output actually produced by the agent, while ``correct action at \(u_t\)'' denotes the ideal action.
Black boxes mark critical error step columns, and diamonds mark ideal memory use.
}
  \label{fig:attention_diagnosis}
\end{figure}

If the context utilization matrix $\mathbf{C}$ truly reflects how memory is used during execution, it should not only assign high scores to useful segments, but also reveal distinct modes of memory use and misuse.
We therefore examine whether the attention-derived matrix $\mathbf{C}$ exposes interpretable utilization patterns at critical decision steps during task execution.
By inspecting context utilization matrix across failed and successful executions, we find that many cases exhibit recurring utilization patterns that consistently correspond to different execution behaviors.
These recurring patterns provide further evidence that attention reflects underlying memory utilization.
We summarize these patterns into three failure scenarios and one successful scenario, each corresponding to a targeted segment-level refinement strategy.
For failed tasks, we first identify the earliest critical error step in \(\tau\) and the segment in \(m^*\) that supports the correct action at that step, which together define the ideal memory use.
We then compare this ideal memory use with the observed utilization in \(\mathbf{C}\) to categorize the failure scenario.
Figure~\ref{fig:attention_diagnosis} presents representative examples, together with their ideal memory use, observed utilization, and corresponding refinement actions.

\noindent\textbf{Attended but wrong.}
This failure scenario occurs when the agent attends to the relevant segment but still makes an incorrect decision. 
In Figure~\ref{fig:attention_diagnosis}(a), the critical error occurs at $u_3$, where the agent should use $s_3$, which contains a \texttt{heat} action, to execute \texttt{heat mug 2 with microwave 1}.
The highlighted $\mathbf{C}_{3,3}$ shows that $s_3$ is indeed accessed. 
However, the agent still outputs the wrong action \texttt{open microwave 1}. 
A closer inspection shows that $s_3$ contains misleading phrasing about opening the microwave, which biases the agent toward an unnecessary \texttt{open} action. 
More broadly, we observe many similar failures caused by misleading or ambiguous content in the attended segment.
We categorize this scenario as \emph{attended but wrong}; the corresponding refinement action is to correct or clarify the attended segment in place, e.g., removing misleading phrasing, fixing incorrect action descriptions, or explicitly distinguishing the correct action from potentially confusing alternatives.


\noindent\textbf{Missed relevant memory.}
This scenario occurs when a relevant memory segment exists but is not attended at the critical error step. 
In Figure~\ref{fig:attention_diagnosis}(b), the agent should use $s_7$ at $u_7$ to perform \texttt{put bread 1 in/on countertop 2}, but instead outputs \texttt{go to countertop 1}. 
The attention pattern shows strong aligned entries at $\mathbf{C}_{5,5}$ and $\mathbf{C}_{6,6}$, but suddenly loses focus at $\mathbf{C}_{7,7}$. 
This indicates that the correct memory $s_7$ is not reliably accessed at the critical step $u_7$, leading to the erroneous action.
We categorize this scenario as \emph{missed relevant memory}; the corresponding refinement action is to emphasize the target segment to improve its salience and retrievability, e.g., by adding explicit task cues, key action keywords, or a clearer instruction-like structure.

\noindent\textbf{Distracted by irrelevant memory.}
This scenario occurs when irrelevant segments attract abnormal attention and interfere with decision making. 
In Figure~\ref{fig:attention_diagnosis}(c), the correct action at $u_2$ should be \texttt{take soapbar 2 from countertop 1}, supported by $s_1$, but the agent outputs \texttt{go to sinkbasin 1}. 
The utilization matrix shows that distractors $s_0$ and $s_2$ receive attention comparable to the relevant segment $s_1$, with $s_2$ even stronger;  moreover, a closer inspection reveals that $s_2$ contains the same action as the erroneous output, while $s_0$ also contains a similar \texttt{go to} pattern. 
This indicates that irrelevant memory interferes with the agent's decision at the critical step $u_2$.
We categorize this scenario as \emph{distracted by irrelevant memory}; the corresponding refinement action is a dual intervention: strengthening the relevant segment with explicit task cues, while rewriting the distracting segment to clarify its scope, reduce irrelevant overlap, or add corrective warnings, thereby turning the original interference source into a guardrail against similar errors.

\noindent\textbf{Low-utilization redundancy.}
In successful executions, $\mathbf{C}$ also provides useful refinement signals. 
Figure~\ref{fig:attention_diagnosis}(d) shows the context utilization matrix of a successful case, where segments $s_{12}$--$s_{18}$ receive little attention across the whole trajectory. 
A closer inspection shows that they mainly contain low-value exploratory actions such as \texttt{go to \{recep\}}. 
We categorize this scenario as \emph{low-utilization redundancy} and treat these weakly utilized segments as candidates for simplification. 
Concretely, we measure low-utilization redundancy by computing the average and maximum utilization of each segment: 
$\bar{c}_\ell=\frac{1}{T+1}\sum_{t=0}^{T}\mathbf{C}_{\ell,t}$ and $c^{\max}_\ell=\max_t\mathbf{C}_{\ell,t}$.
A segment is considered weakly utilized only when both scores fall below predefined thresholds.

Together, these patterns show that attention reveals distinct states of memory utilization, including used-but-misleading, missed, distracting, and weakly utilized memory.

\section{Attention-Guided Memory Refinement}
\label{sec:method}

\begin{figure}[t]  
  \centering
  \includegraphics[width=\linewidth]{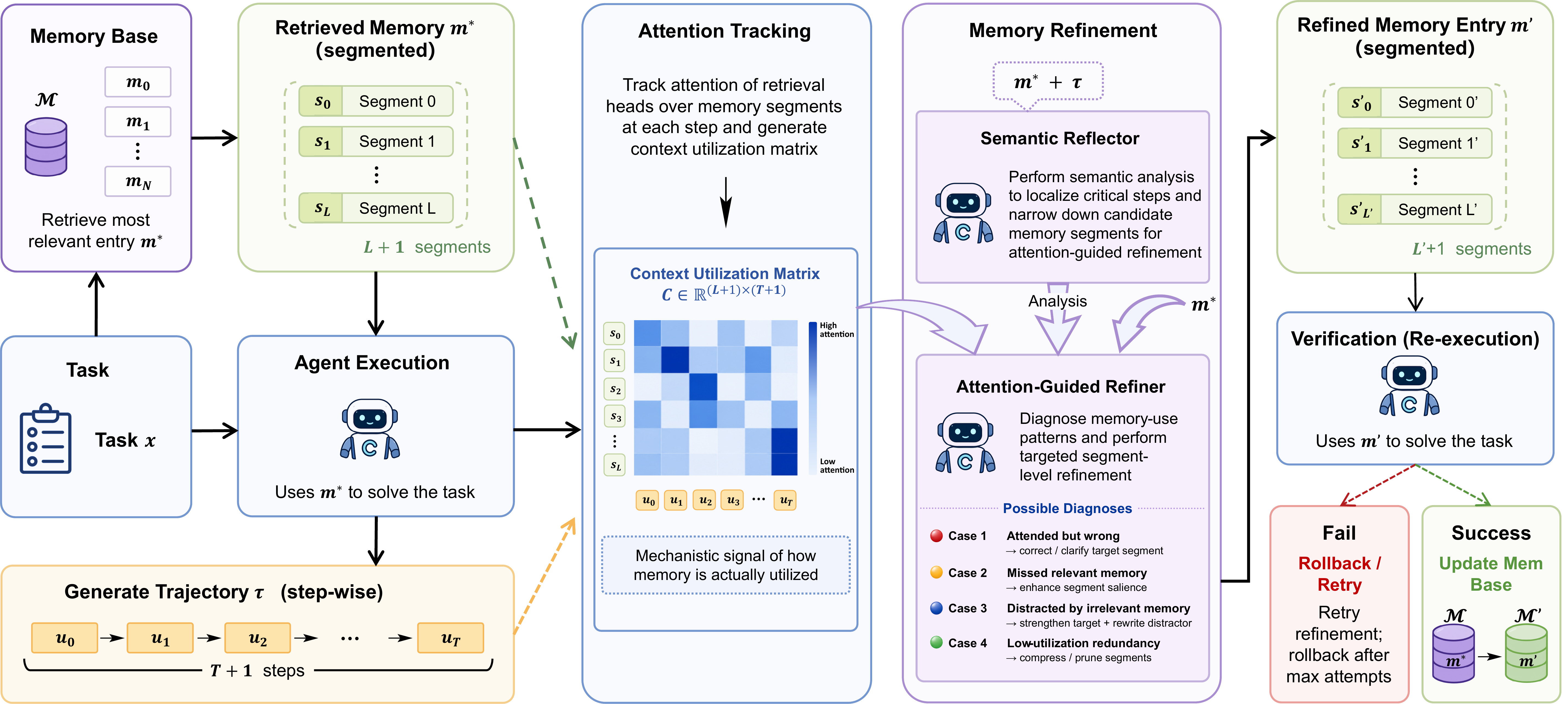}\vspace{-6pt}\caption{
Overview of AGMR. AGMR retrieves a segmented trajectory memory entry for a task,
tracks retrieval-head attention during execution to construct a context utilization matrix, and uses this signal to guide targeted segment-level memory refinement. Refined memories are committed only after successful re-execution verification.
}
\vspace{-3pt}
  \label{fig:main}
\end{figure}

Building on the attention-revealed utilization patterns identified in Section~\ref{sec:different_patterns}, we propose Attention-Guided Memory Refinement (AGMR), a framework that converts retrieval-head attention into structured evidence of memory utilization and uses this mechanistic guidance to drive targeted segment-level memory refinement.
Figure~\ref{fig:main} illustrates the overall pipeline of AGMR. 
Given a task \(x\) with task description \(d_x\), the agent retrieves a memory entry \(m^*\) from the memory base \(\mathcal{M}\) by selecting the entry whose associated task description has the highest cosine similarity to \(d_x\) in the embedding space.
The retrieved memory is segmented as \(m^*=[s_0,\ldots,s_L]\) and injected into the prompt to support step-wise task execution, producing a trajectory \(\tau=[u_0,\ldots,u_T]\).

During execution, AGMR tracks retrieval-head attention over memory segments at each
decision step. By aggregating these signals across selected retrieval heads and stacking them over time, AGMR constructs a context utilization matrix \(\mathbf{C}\in\mathbb{R}^{(L+1)\times(T+1)}\), where each entry indicates how strongly a memory segment is accessed before a particular execution step.

AGMR then performs two-stage memory refinement according to the task outcome. 
For failed executions, the Semantic Reflector first localizes the critical error and candidate useful memory segment, and the Attention-Guided Refiner then uses the context utilization matrix \(\mathbf{C}\) to guide targeted correction or enhancement. 
For successful executions, the Semantic Reflector identifies semantically redundant candidates, and the Attention-Guided Refiner uses utilization evidence in \(\mathbf{C}\) to decide which segments can be safely simplified. 
Each candidate update is verified by re-executing the task with the refined memory \(m'\). 
Only updates that pass verification are committed to the memory base; otherwise, AGMR rolls back or retries the refinement.

\subsection{Two-stage memory refinement}
\label{sec:memory_refinement}

Given a memory entry \(m^*\), its task trajectory \(\tau\), and the context utilization matrix \(\mathbf{C}\), AGMR refines memory through two cooperative LLM stages: a \emph{Semantic Reflector} and an \emph{Attention-Guided Refiner}.
The Semantic Reflector narrows the analysis scope through semantic reasoning over the memory and trajectory, while the Attention-Guided Refiner uses \(\mathbf{C}\) to identify the memory-use pattern and produce targeted segment-level updates.
Moreover, the same two stages serve different functions depending on whether the execution fails or succeeds.
The prompts are provided in Appendix~\ref{app:prompt_details}.

\noindent\textbf{Failure cases.}
For a failed execution, the Semantic Reflector analyzes the failed trajectory \(\tau\) together with the retrieved memory \(m^*\) and produces a structured analysis $A_f$. 
This analysis localizes the earliest critical error step $u_j$, infers the correct action that should have been taken at $u_j$, and identifies the memory segment $s_i$ that supports this correct action. 
Although this textual analysis localizes the error, it does not determine the underlying memory-use mechanism that drove the wrong decision.
AGMR therefore passes $A_f$, the retrieved memory \(m^*\), and the erroneous-step utilization column $\mathbf{C}_{:,j}$ to the Attention-Guided Refiner. 
The Attention-Guided Refiner compares the ideal memory use $u_j\rightarrow s_i$ with the observed utilization column $\mathbf{C}_{:,j}$ and determines whether the failure arises from \emph{attended but wrong}, \emph{missed relevant memory}, or \emph{distracted by irrelevant memory}. 
It then applies the corresponding update: in-place correction of \(s_i\), salience-oriented enhancement of \(s_i\), or dual modification of \(s_i\) and the distracting segment, producing a refined memory \(m'\). 
Thus, AGMR localizes the visible error with textual analysis, but grounds memory-use attribution and targeted segment-level refinement in mechanistic attention evidence.

\noindent\textbf{Success cases.}
For a successful execution, AGMR performs memory refinement for simplification rather than correction. 
The Semantic Reflector first compares the successful trajectory \(\tau\) with the retrieved memory \(m^*\) to identify semantically redundant segments, producing a candidate list \(\mathcal{R}\).
Since semantic redundancy alone does not guarantee that a segment is safe to remove, the Attention-Guided Refiner further evaluates each candidate using utilization statistics from \(\mathbf{C}\). 
For each segment \(s_\ell\in\mathcal{R}\), we compute its average and maximum utilization, \(\bar{c}_\ell = \frac{1}{T+1}\sum_{t=0}^{T}\mathbf{C}_{\ell,t}\) and \(c^{\max}_\ell = \max_t \mathbf{C}_{\ell,t}\). 
A segment is simplified only if it is both semantically redundant (i.e., in \(\mathcal{R}\)) and weakly utilized (both \(\bar{c}_\ell\) and \(c^{\max}_\ell\) fall below predefined thresholds). 
The Refiner then deletes or compresses the qualified segments, and minimally adjusts neighboring content to preserve local coherence, producing the refined memory \(m'\).

\subsection{Re-execution verification}
\label{sec:reexecution_verification}

AGMR does not immediately commit a candidate memory update. 
Instead, the refined memory \(m'\) is validated by re-executing the task using \(m'\). 
The commit policy depends on the branch.
For the failure branch, a refinement is accepted if it enables successful task completion or achieves a performance improvement exceeding a predefined threshold.
If re-execution still fails, AGMR retries the refinement using \(m'\), the new trajectory and utilization matrix, up to a fixed maximum number of attempts. 
If all attempts fail, the update is discarded and the memory is rolled back to the original \(m^*\).
For the success branch, the goal is to simplify memory without harming task success. 
Therefore, AGMR adopts a stricter policy: if re-execution preserves success, the simplified memory is committed; otherwise, the update is immediately rolled back without further retry. 
This verification step prevents harmful edits from accumulating and grounds memory updates in actual task performance.

\section{Experiments}
\label{sec:experiments}

\begin{table*}[t]
\centering
\small
\caption{Performance comparison across different environments and models.}
\label{tab:main_results}

\setlength{\tabcolsep}{4pt}
\renewcommand{\arraystretch}{1.08}

\begin{tabularx}{\textwidth}{
>{\centering\arraybackslash}p{0.13\textwidth}
>{\centering\arraybackslash}p{0.09\textwidth}
>{\centering\arraybackslash}p{0.12\textwidth}
>{\centering\arraybackslash}X
>{\centering\arraybackslash}X
>{\centering\arraybackslash}X
>{\centering\arraybackslash}X
>{\centering\arraybackslash}X
>{\centering\arraybackslash}X
}
\toprule
\textbf{Model} & \textbf{Env} & \textbf{Metric}
& \textbf{ReAct} & \textbf{ICL} & \textbf{AWM}
& \textbf{DC} & \textbf{Traj} & \textbf{AGMR} \\
\midrule

\multirow{8}{*}{\textbf{\makecell{Qwen-2.5\\7B-Instruct}}}
& \multirow{3}{*}{ALFWorld}
& Perf. $\uparrow$
& 41.79 & 45.52 & 64.18 & 52.24 & \underline{70.90} & \textbf{81.34} \\
& & Avg step $\downarrow$
& 28.40 & 26.71 & 22.22 & 25.09 & \underline{19.36} & \textbf{16.08} \\
& & Mem token $\downarrow$
& -- & -- & \textbf{312.01} & 1013.73 & 852.36 & \underline{752.53} \\

\cmidrule(lr){2-9}

& \multirow{3}{*}{SciWorld}
& Perf. $\uparrow$
& 29.88 & 29.38 & 50.39 & 63.73 & \underline{67.69} & \textbf{77.20} \\
& & Avg step $\downarrow$
& 19.56 & 18.35 & 16.92 & 15.52 & \underline{14.62} & \textbf{13.78} \\
& & Mem token $\downarrow$
& -- & -- & \textbf{435.32} & 1478.85 & 1040.62 & \underline{990.89} \\

\cmidrule(lr){2-9}

& \multirow{2}{*}{WebShop}
& Perf. $\uparrow$
& 56.66 & \underline{60.12} & 56.11 & 54.04 & 51.04 & \textbf{63.86} \\
& & Mem token $\downarrow$
& -- & -- & \textbf{185.65} & 618.14 & 1372.22 & \underline{503.77} \\

\midrule

\multirow{8}{*}{\textbf{\makecell{Llama-3.1\\8B-Instruct}}}
& \multirow{3}{*}{ALFWorld}
& Perf. $\uparrow$
& 0.00 & 1.49 & 7.46 & 11.19 & \underline{26.12} & \textbf{47.76} \\
& & Avg step $\downarrow$
& 40.00 & 39.74 & 38.30 & 36.67 & \underline{32.54} & \textbf{25.52} \\
& & Mem token $\downarrow$
& -- & -- & \textbf{312.01} & 1133.51 & 852.36 & \underline{845.91} \\

\cmidrule(lr){2-9}

& \multirow{3}{*}{SciWorld}
& Perf. $\uparrow$
& 26.50 & 34.97 & 52.14 & 72.28 & \underline{76.18} & \textbf{83.74} \\
& & Avg step $\downarrow$
& 19.59 & 18.37 & 16.23 & 13.99 & \underline{13.22} & \textbf{12.04} \\
& & Mem token $\downarrow$
& -- & -- & \textbf{435.32} & 1168.99 & 1040.62 & \underline{918.53} \\

\cmidrule(lr){2-9}

& \multirow{2}{*}{WebShop}
& Perf. $\uparrow$
& 33.06 & \underline{52.10} & 41.56 & 29.68 & 41.45 & \textbf{63.26} \\
& & Mem token $\downarrow$
& -- & -- & \textbf{185.65} & 658.96 & 1372.22 & \underline{567.57} \\

\bottomrule
\end{tabularx}
\label{tab:main_results}
\end{table*}

\subsection{Experimental setup}
We evaluate AGMR on three interactive decision-making benchmarks: ALFWorld~\cite{shridhar2020alfworld}, SciWorld~\cite{wang2022scienceworld}, and WebShop~\cite{yao2022webshop}, covering embodied household interaction, science-oriented multi-step reasoning, and web-based decision making. 
We use Qwen2.5-7B-Instruct~\cite{qwen2025qwen25technicalreport} and Llama-3.1-8B-Instruct~\cite{grattafiori2024llama3herdmodels} as the backbone models for the agent. 
Each task is solved by retrieving a memory entry from the memory base and injecting it into the prompt. 
We report Performance on all benchmarks, Average Step on ALFWorld and SciWorld, and Average Token Number of Memory Entries on all benchmarks. 
Performance measures task reward, Avg step measures execution efficiency, and Memory token measures the prompt length contributed by retrieved memory. 
We omit Avg step for WebShop because achieving high reward requires purchasing a product that satisfies multiple attributes, which inevitably involves additional attribute inspection and clicks. 
Detailed dataset configurations and hyperparameter settings, including those used in Section~\ref{sec:memory_refinement}, are provided in Appendix~\ref{app:experiments_details}.

Baselines include ReAct~\cite{yao2022react}, the base agent without memory; 
ICL, a fixed in-context demonstration; 
AWM~\cite{wang2024awm}, an agent workflow memory method that extracts reusable workflow abstractions from past trajectories; 
DC~\cite{suzgun2026dynamic}, a dynamic cheatsheet method that adapts memory from prior experience; 
and Traj, which directly inserts raw retrieved trajectory memory extracted from the SFT dataset released by ETO~\cite{song2024trial}, without refinement.
For AWM and DC, we use the offline version of AWM and the retrieval-based DC-RS variant of DC, respectively.
AGMR uses the raw trajectory memory from Traj as its initial memory base and refines it, so comparison with Traj isolates the effect of our memory refinement.
AGMR refines the initial memory base solely using the training set, producing an refined memory base that is directly used for evaluation on the test set without additional modification.
For fair comparison, the \emph{Semantic Reflector}, \emph{Attention-Guided Refiner}, AWM workflow extraction, and Dynamic Cheatsheet's \emph{Memory Curator} all use Gemini-3.1-pro-preview with temperature 1.0 and high thinking mode.

\subsection{Main results}

\begin{figure}[t]  
  \centering
  \includegraphics[width=\linewidth]{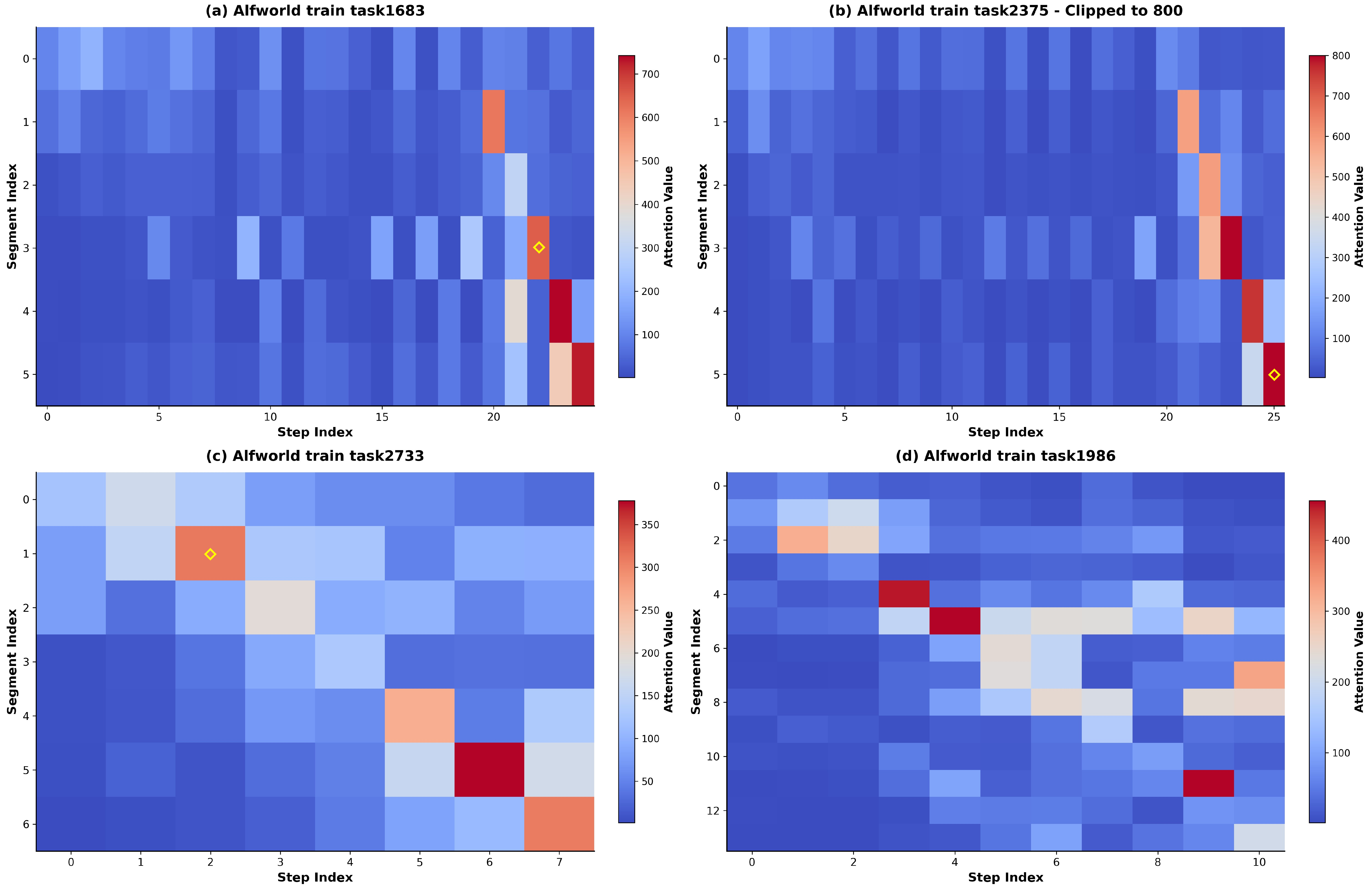}\vspace{-6pt}\caption{Context utilization matrix after AGMR; diamonds denote expected memory use.}
  \label{fig:attention_after_refine}
  \vspace{-3pt}
\end{figure}

Table~\ref{tab:main_results} reports the main results of AGMR and all baselines. 
Overall, AGMR achieves the best performance across all three environments and both backbone models. 
It also obtains the best execution efficiency on ALFWorld and SciWorld, where Avg step is reported, indicating that refined memory helps agents complete tasks more directly with fewer unnecessary actions. 

\begin{wrapfigure}[13]{r}{0.38\linewidth}  
  \centering
  \vspace{-10pt}
  \includegraphics[width=\linewidth]{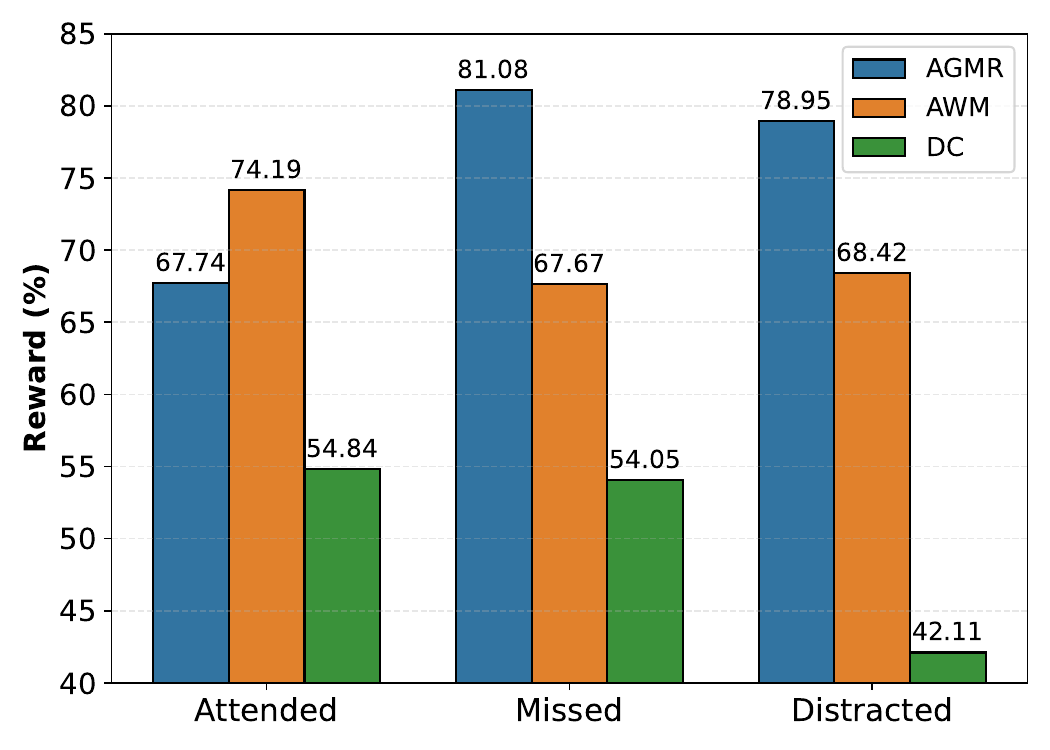}\vspace{-6pt}\caption{
Scenario-wise performance comparison.
}
  \label{fig:scenario_comparison}
\end{wrapfigure}

The most direct comparison is between AGMR and Traj, since AGMR is refined from the trajectory memory. 
Across all settings, AGMR improves performance over Traj while generally reducing the number of memory tokens. 
This suggests that AGMR does not simply benefit from adding more context; instead, it improves memory quality by strengthening useful segments and removing or compressing low-value content, realizing memory \emph{refinement} rather than mere memory expansion.


AGMR also outperforms strong memory baselines such as AWM and DC. 
To better understand the source of this improvement, we further analyze the Qwen-ALFWorld refinement process by grouping memory entries according to the three failure scenarios reported by the Attention-Guided Refiner, and compute the performance of tasks associated with each memory-entry group.
As shown in Figure~\ref{fig:scenario_comparison}, AGMR's advantage mainly comes from Scenario~2 (\emph{Missed relevant memory}) and Scenario~3 (\emph{Distracted by irrelevant memory}), especially the latter, while AWM performs better in Scenario~1 (\emph{Attended but wrong}). 
This is consistent with our motivation: Scenario~1 often involves misleading or ambiguous memory content, which can be identified and mitigated by text-based analysis, abstraction, and summarization. 
In contrast, Scenario~2 and Scenario~3 depend on how memory is actually utilized during execution: whether the relevant segment is missed, or whether irrelevant segments compete for attention.
Such utilization patterns are difficult to infer from textual trajectories alone, but become visible in the context utilization matrix.
AGMR can therefore address these failures through targeted memory refinement.
These results support our central claim that attention reveals memory utilization and provides actionable guidance for memory refinement.

\subsection{How AGMR reshapes memory utilization}


\begin{wraptable}[13]{r}{0.43\linewidth}
\vspace{-16pt}
\centering
\scriptsize
\setlength{\tabcolsep}{3pt}
\renewcommand{\arraystretch}{0.92}
\caption{
Ablation of the Attention-Guided Refiner.
\textit{w/o AGR} lets the Semantic Reflector directly edit memory.
}
\label{tab:ablation_ad}
\begin{tabular}{llcc}
\toprule
\textbf{Setting} & \textbf{Metric} & \textbf{w/o AGR} & \textbf{AGMR} \\
\midrule
Qwen-ALF & Perf. $\uparrow$ & 77.61 & \textbf{81.34} \\
Qwen-ALF & Step $\downarrow$ & 18.13 & \textbf{16.08} \\
Qwen-Sci & Perf. $\uparrow$ & 67.98 & \textbf{77.20} \\
Qwen-Sci & Step $\downarrow$ & 14.77 & \textbf{13.78} \\
Qwen-Web & Perf. $\uparrow$ & 57.76 & \textbf{63.86} \\
\midrule
Llama-ALF & Perf. $\uparrow$ & 37.31 & \textbf{47.76} \\
Llama-ALF & Step $\downarrow$ & 29.25 & \textbf{25.52} \\
Llama-Sci & Perf. $\uparrow$ & 80.33 & \textbf{83.74} \\
Llama-Sci & Step $\downarrow$ & 12.56 & \textbf{12.04} \\
Llama-Web & Perf. $\uparrow$ & 54.90 & \textbf{63.26} \\
\bottomrule
\end{tabular}
\vspace{-8pt}
\end{wraptable}

To better understand how AGMR changes memory use in practice, we compare pre- and post-refinement context utilization matrices for the representative cases in Figure~\ref{fig:attention_diagnosis}; the post-refinement matrices are shown in Figure~\ref{fig:attention_after_refine}.
The results show that refined memories induce the expected step-segment utilization patterns. 
Note that the post-refinement critical step may not share the same index as before refinement, since prompt changes can alter the agent's exploratory path before it reaches the same key decision.

For the attended but wrong case, after AGMR corrects the misleading content in \(s_3\), Figure~\ref{fig:attention_after_refine}(a) shows that the model still attends to the relevant segment \(s_3\) at the decisive step, now \(u_{22}\) in the re-executed trajectory. 
Unlike before refinement, this attention now supports the correct \texttt{heat} action rather than the erroneous \texttt{open} action, and the agent completes the task two steps later. 
For the missed relevant memory case, after AGMR rewrites the target segment to improve its salience, Figure~\ref{fig:attention_after_refine}(b) shows that attention is pulled back to the correct memory segment. 
Due to simplification, the original \(s_7\) becomes \(s_5\) in the refined memory; at \(u_{25}\), the highlighted \(\mathbf{C}_{5,25}\) indicates that the agent attends to this segment, uses it to produce the correct action, and completes the task successfully. 
For the distracted by irrelevant memory case, after AGMR strengthens the relevant segment and adds clarifying information to the distractor, Figure~\ref{fig:attention_after_refine}(c) shows that the utilization pattern is corrected. 
Attention shifts toward the target segment \(s_1\), with \(\mathbf{C}_{1,2}\) highlighted, while the former distractor \(s_2\) no longer shows significant interference at the critical step. 
Finally, Figure~\ref{fig:attention_after_refine}(d) illustrates utilization-aware pruning: after AGMR removes weakly utilized redundant content, the memory shrinks from 23 to 13 segments, and the resulting trajectory becomes visibly denser, with far fewer segments remaining entirely unused.
Overall, the post-refinement context utilization matrices show that AGMR does more than rewrite memory text: it reshapes memory utilization in ways that match the diagnosed memory-use pattern.

\subsection{Ablation study}
\label{sec:ablation}





To assess the role of mechanistic attention guidance, we conduct an ablation study by removing the Attention-Guided Refiner.
In this variant, the Semantic Reflector directly edits memory segments after producing the structured analysis \(A_f\).
As shown in Table~\ref{tab:ablation_ad}, removing the Attention-Guided Refiner consistently degrades performance, indicating the importance of attention-guided diagnosis.

\section{Conclusion}
\label{sec:conclusion}
We study how retrieved memory is utilized during agent execution and showed that retrieval-head attention can reveal segment-level memory-use patterns through a context utilization matrix. 
Based on this observation, we proposed AGMR, which uses these patterns to guide targeted memory refinement. 
Experiments show that AGMR improves both task performance and memory efficiency over text-only memory refinement baselines. 

\noindent\textbf{Limitations} This work is currently limited to trajectory memory, where memory can be naturally segmented into discrete interaction steps. Consequently, the current AGMR code may not directly generalize to other forms of memory. 
Another limitation is that the refinement pipeline is performed offline before evaluation. Incorporating AGMR into an online or real-time update setting is left for future exploration.

\noindent\textbf{Further work} Future work will extend AGMR to other memory forms, such as workflow or procedural memory, and explore online refinement, where trajectories generated during test-time execution are dynamically incorporated into the memory base for continual agent improvement.

\clearpage
\bibliographystyle{plain}
\bibliography{references}

\clearpage
\appendix

\section{Related works}

\paragraph{Memory refinement in LLM agents.}
Memory has become a central component of LLM agents, enabling past interactions to be stored and reused for long-horizon decision making. 
Prior work has explored diverse memory forms, including trajectory exemplars, contextual memories for planning, structured long-term memory, and hierarchical memory management~\cite{zheng2023synapse, kagaya2024rap, packer2023memgpt, hu2024hiagent, xu2025mem, hu2025memory}. 
More recent systems further treat memory as an evolving component that can be updated through reflection, adaptive refinement, or continual experience accumulation~\cite{shinn2023reflexion, suzgun2026dynamic, zhang2025ace, cai2025flex}. 
However, these methods typically rely on model-generated trajectories or textual analyses as the primary update signal. 
Since LLM-generated explanations can be plausible yet unfaithful~\cite{turpin2023language, agarwal2024faithfulness, chuang2026faithlm, arcuschin2025cotwild}, textual reflection alone may provide a weak basis for grounded memory refinement. 
AGMR instead studies how internal mechanistic signals can guide targeted updates to specific memory content.

\paragraph{Attention and mechanistic interpretability.}
Mechanistic interpretability studies suggest that attention heads can specialize in accessing and routing contextual information. 
Most relevantly, prior work identifies sparse retrieval heads that selectively attend to key information for long-context prediction and retrieve relevant context from prompts~\cite{wu2025retrievalheads, kahardipraja2025atlasicl}. 
Related work further finds query-dependent heads and gathering or routing heads that support contextual information selection and integration~\cite{zhang2025query, bick2025gather}. 
Together, these studies suggest that retrieval-related attention provides useful mechanistic signals for analyzing how contextual information is accessed during generation. 
AGMR builds on this perspective by aggregating retrieval-head attention over memory segments into a context utilization matrix, using it as structured evidence for segment-level memory diagnosis.

\paragraph{Attention-based context evolution.}
A related line of work uses attention to identify and reshape important context for efficient long-context inference. 
AttnComp exploits attention to guide prompt compression~\cite{zhao2025attncomp}; EHPC identifies evaluator heads that retain key tokens in long inputs~\cite{fei2025efficient}; and QUITO applies query-guided attention to filter irrelevant context~\cite{wang2024quito}. 
KV-cache methods such as H$_2$O, Scissorhands, and SnapKV similarly use attention-derived importance to retain high-value tokens or cache entries~\cite{zhang2023h2o, liu2023scissorhands, li2024snapkv}. 
These methods share AGMR's view that attention reveals contextual importance, but they primarily optimize inference efficiency by compressing prompts or caches. 
In contrast, AGMR focuses on agent memory refinement: it diagnoses missed, misleading, distracting, and redundant memory-use patterns, then updates memory content rather than merely removing low-score context.

\section{Algorithm of AGMR}
\label{app:agmr_algorithm}

\begin{algorithm}[t]
\caption{Attention-Guided Memory Refinement (AGMR)}
\label{alg:agmr}
\begin{algorithmic}[1]
\REQUIRE Task \(x\), memory base \(\mathcal{M}\), maximum retry number \(K\)
\ENSURE Updated memory base \(\mathcal{M}\)

\STATE Retrieve the most relevant trajectory memory \(m^*\) from \(\mathcal{M}\) for task \(x\)
\STATE Execute the agent with \(m^*\), obtaining trajectory \(\tau\), task outcome \(o\), and context utilization matrix \(\mathbf{C}\)

\IF{\(o\) is failure}
    \STATE Semantic Reflector analyzes \((m^*, \tau)\) and produces structured analysis \(A_f\)
    \STATE Attention-Guided Refiner uses \((A_f, m^*, \mathbf{C})\) to diagnose the memory-use pattern
    \STATE Generate a refined memory \(m'\) by correcting, enhancing, or modifying target segments
    \STATE Verify \(m'\) by re-executing task \(x\)
    \STATE Retry refinement up to \(K\) times if verification fails
\ELSE
    \STATE Semantic Reflector identifies semantically redundant candidates \(\mathcal{R}\) from \((m^*, \tau)\)
    \STATE Attention-Guided Refiner filters \(\mathcal{R}\) using utilization statistics from \(\mathbf{C}\)
    \STATE Generate a simplified memory \(m'\) by compressing or removing redundant weakly utilized segments
    \STATE Verify \(m'\) by re-executing task \(x\)
\ENDIF

\IF{verification succeeds}
    \STATE Commit \(m'\) to \(\mathcal{M}\)
\ELSE
    \STATE Discard \(m'\) and keep the original memory \(m^*\)
\ENDIF

\RETURN \(\mathcal{M}\)
\end{algorithmic}
\end{algorithm}

Algorithm~\ref{alg:agmr} summarizes the overall workflow of Attention-Guided Memory Refinement (AGMR). 
Given a task and a memory base, AGMR first retrieves the most relevant trajectory memory and executes the agent to obtain a task trajectory along with the context utilization matrix derived from attention. 
Depending on whether the execution succeeds or fails, the framework applies a two-stage refinement process: the \emph{Semantic Reflector} identifies candidate memory segments for correction, enhancement, or simplification, and the \emph{Attention-Guided Refiner} uses attention-derived utilization signals to guide targeted segment-level updates. 
Each candidate memory update is validated via re-execution before being committed to the memory base, ensuring that only verified improvements are incorporated. 
This algorithm provides a high-level overview of AGMR, complementing the detailed description in Section~\ref{sec:method}.

\section{Retrieval head identification}
\subsection{Construction of retrieval supervision dataset}
\label{app:retrieval_dataset}

To identify retrieval heads, we construct a supervision dataset of 
$(\texttt{key\_context}, \texttt{formatted\_prompt})$ pairs for each backbone model. 
The goal is to collect cases where a specific memory snippet is known to be useful for a key decision step, so that attention heads can be scored by whether they attend to this snippet before the step is generated.

We build this dataset from tasks whose outcomes are improved by adding retrieved trajectory memory. 
Taking Qwen-2.5-7B-Instruct on ALFWorld as an example, the agent achieves a reward of 0.4179 without memory. 
After inserting the most similar retrieved memory trajectory into the prompt, the reward increases to 70.90\%. 
From this setting, we collect tasks that fail without memory but succeed with memory. 
These transition cases provide natural supervision, since the retrieved memory is likely to contain information that helps correct the original failure.

For each such task, we use an LLM analyzer to compare the original failed trajectory with the subsequent successful memory-augmented trajectory. 
The analyzer identifies the failure reason, the key step in the successful trajectory that contributes to the failure-to-success transition, and the corresponding memory snippet that supports this key step. 
This supporting snippet is recorded as \texttt{key\_context}. 

Based on this analysis, each supervision example is stored as a tuple 
$(\texttt{task\_id}, \texttt{key\_context}, \texttt{formatted\_prompt})$. 
Here, \texttt{formatted\_prompt} denotes the complete model input immediately before generating the identified key step, and \texttt{key\_context} denotes the memory snippet expected to be retrieved at that moment. 
The same construction procedure is applied to Llama-3.1-8B-Instruct, producing model-specific retrieval supervision datasets for head scoring.

\subsection{Retrieval score and head distribution}
\label{app:retrieval_score}

\begin{figure}[t]  
  \centering
  \includegraphics[width=\linewidth]{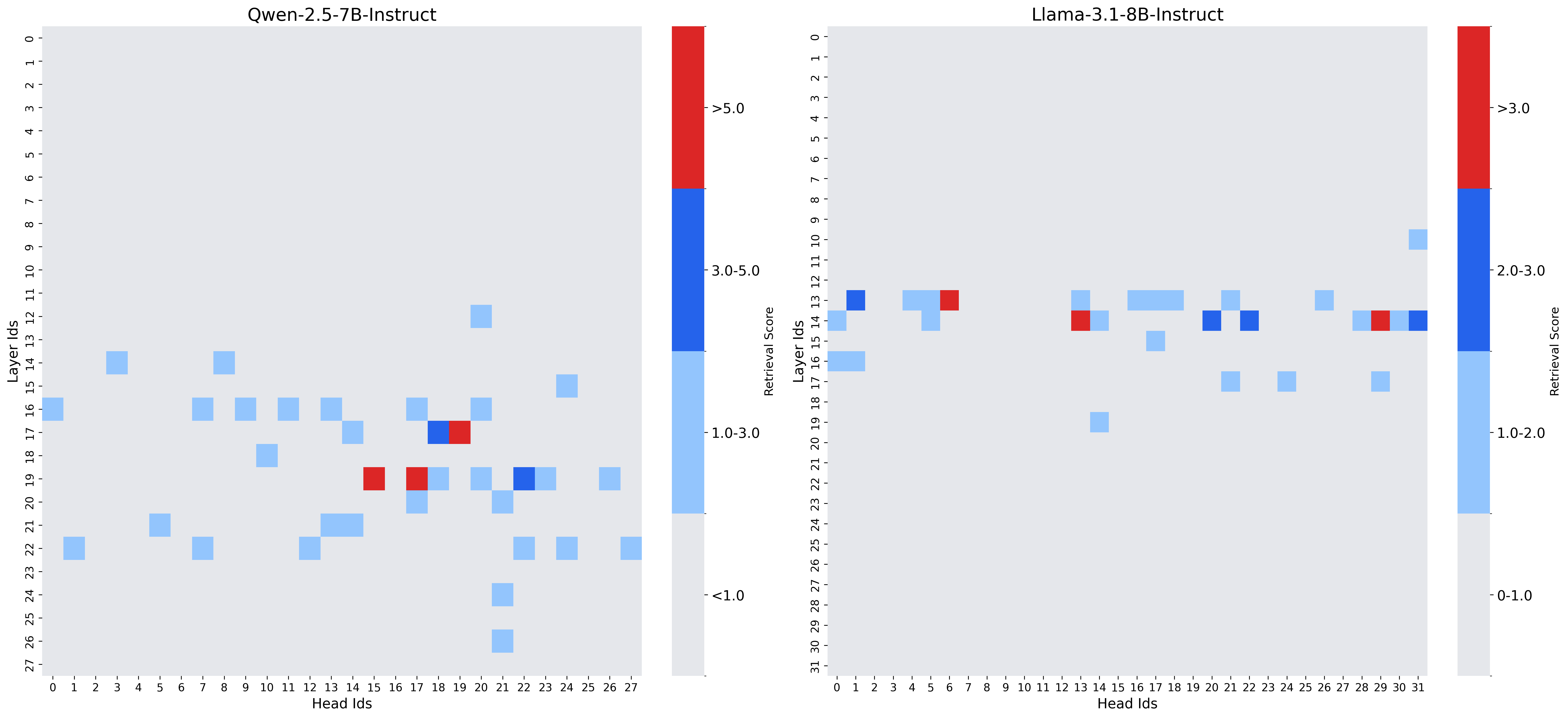}
\caption{
Retrieval Score heatmaps for all attention heads in Qwen-2.5-7B-Instruct (left) and Llama-3.1-8B-Instruct (right). Each entry shows how much attention a head assigns from the prompt's last token to the annotated \texttt{key\_context} across the retrieval supervision set. Only a small subset of heads exhibits markedly high Retrieval Scores, and these heads are concentrated in middle layers, suggesting localized retrieval-related functionality.
}
  \label{fig:retrieval_score_heatmap}
\end{figure}

Given the retrieval supervision dataset, we compute a Retrieval Score for every attention head. 
For a supervision pair $(\texttt{key\_context}, \texttt{formatted\_prompt})$, we examine the attention distribution from the last token of \texttt{formatted\_prompt} to all previous context tokens. 
For each layer $l$ and head $h$, we define the task-level retrieval score as:
\begin{equation}
\mathrm{score}_{l,h}^{(\mathrm{task})}
=
\sum_{t \in \mathrm{key\_context\_tokens}}
\mathrm{Attention}_{l,h}(\mathrm{last\_token} \rightarrow t),
\end{equation}
where $\mathrm{Attention}_{l,h}(\mathrm{last\_token} \rightarrow t)$ denotes the attention weight from the last token to token $t$ in the annotated key context.

We then aggregate this value over all supervision pairs:
\begin{equation}
\mathrm{score}_{l,h}
=
\sum_{\mathrm{task}}
\mathrm{score}_{l,h}^{(\mathrm{task})}.
\end{equation}
A high Retrieval Score indicates that the head consistently assigns strong attention to memory snippets that are useful for the next decision step.

Figure~\ref{fig:retrieval_score_heatmap} visualizes the Retrieval Score distributions for Qwen-2.5-7B-Instruct and Llama-3.1-8B-Instruct. 
In both models, only a small subset of heads obtains markedly high scores, suggesting that memory retrieval behavior is concentrated in a limited number of specialized heads rather than uniformly distributed across all heads.

We select the top-5 heads with the highest retrieval scores as the retrieval-head set \(\mathcal{H}_{\mathrm{ret}}\). 
Specifically, for Qwen2.5-7B-Instruct, we select \((17,18)\), \((17,19)\), \((19,15)\), \((19,17)\), and \((19,22)\); for Llama-3.1-8B-Instruct, we select \((13,6)\), \((14,13)\), \((14,29)\), \((14,31)\), and \((13,1)\), where each pair denotes \((\mathrm{Layer}, \mathrm{Head})\).

\subsection{Ablation validation of retrieval heads}
\label{app:retrieval_head_ablation}

To verify that the identified high-score heads are functionally important for memory-assisted behavior, we conduct an attention-head masking experiment. 
Evaluation is performed under the raw trajectory-memory setting, where the agent receives retrieved trajectory memory in the prompt. 
For Qwen-2.5-7B-Instruct, we mask the top 10 heads ranked by Retrieval Score and compare them with 10 randomly selected heads. 
For Llama-3.1-8B-Instruct, we analogously mask the top 5 retrieval heads and 5 random heads.

As shown in Figure~\ref{fig:retrieval_head_ablation}, masking the high-Retrieval-Score heads substantially reduces the performance gains brought by trajectory memory. 
In contrast, masking randomly selected heads has little effect. 
This result indicates that the selected high-score heads are not merely correlated with useful memory snippets, but play an important functional role in extracting relevant information from memory. 
Therefore, we use the top 5 retrieval heads of each backbone model as the basis for context utilization tracking in AGMR.

\section{Distribution of different memory-utilization patterns}
\label{app:utilization_pattern_distribution}
\begin{figure}[t]  
  \centering
  \includegraphics[width=\linewidth]{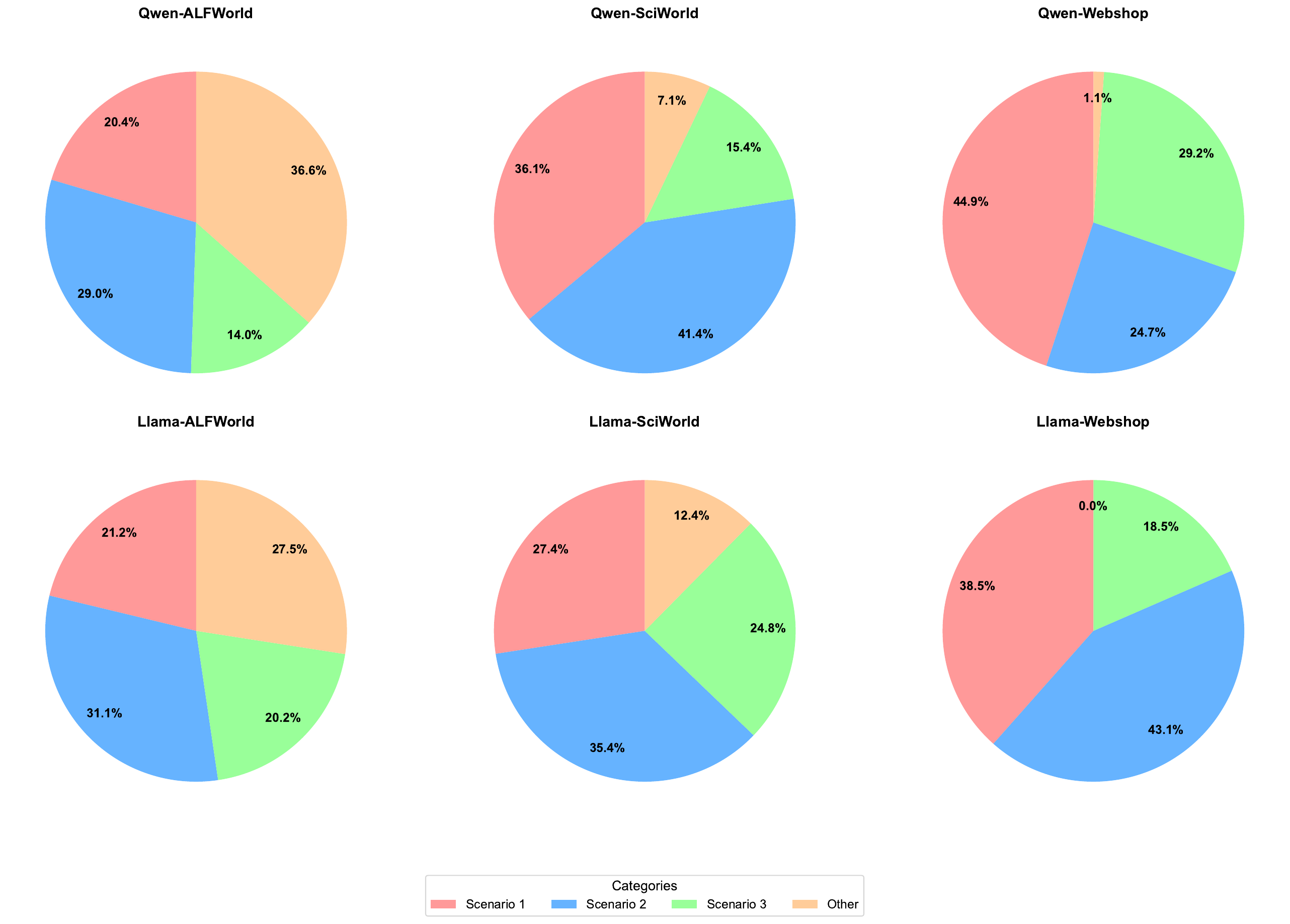}
\caption{
Proportions of memory-use scenarios among failed tasks under six experimental settings.
For each failed task, the reported category is determined by the diagnosis from the Attention-Guided Refiner in the final refinement iteration, accounting for the retry mechanism.
}
  \label{fig:distribution_pie_charts}
\end{figure}

Figure~\ref{fig:distribution_pie_charts} presents the distribution of the three memory-use scenarios among failed tasks across six experimental settings. Given the presence of the retry mechanism, for each failed task, we use the diagnostic result generated by the Attention-Guided Refiner in the final refinement process for statistical analysis.

\section{Experiments details}
\label{app:experiments_details}

\subsection{SciWorld configuration}

For SciWorld, we adopt a filtered version of the original test set due to an implementation constraint in our early attention collection pipeline. 
AGMR requires access to attention weights during model forward passes to construct the context utilization matrix. 
In our initial implementation, the attention-hook based collection method caused incorrect model forward behavior due to a bug. 
Therefore, we instead directly read and stored \texttt{output.attentions} for context utilization matrix construction. 
This fallback implementation introduced substantial GPU memory overhead, especially for task types with large \texttt{max\_steps}, where the growing prompt could lead to out-of-memory errors during execution. 
To avoid such failures, we excluded SciWorld task types with \texttt{max\_steps} $\geq 60$ from the experimental scope, including:

\begin{itemize}[leftmargin=*]
    \item \texttt{task-1-boil},
    \item \texttt{task-1-change-the-state-of-matter-of},
    \item \texttt{task-1-freeze},
    \item \texttt{task-1-melt},
    \item \texttt{task-10-measure-melting-point-(known-substance)},
    \item \texttt{task-4-grow-fruit},
    \item \texttt{task-5-chemistry-mix}.
\end{itemize}

We later fixed the attention-hook implementation and enabled more memory-efficient context utilization tracking, but kept this SciWorld configuration unchanged for consistency with the completed experiments. 
The same filtered subset is used for AGMR and all baselines. 
Thus, this filtering is an early implementation constraint.

\subsection{Hyperparameters for memory refinement}
\label{app:refinement_hyperparameters}

We use benchmark-specific hyperparameters for utilization-aware simplification and re-execution verification. 
For successful-task simplification, a segment is considered weakly utilized according to the attention statistics defined in Section~\ref{sec:memory_refinement}. 
For ALFWorld, we use a relative criterion based on each memory entry: a segment is treated as weakly utilized if
\[
c^{\max}_\ell < 1.2 \cdot \mathrm{median}_{k}(c^{\max}_k)
\quad \text{and} \quad
\frac{c^{\max}_\ell}{\bar{c}_\ell} < 3.5 .
\]
This criterion removes segments whose peak attention is below the entry-level median scale and whose utilization is not concentrated around any specific decision step.

For SciWorld and WebShop, we use a global low-attention threshold. 
In SciWorld, we set
\[
\theta_{\mathrm{low}} = 0.25 \cdot c^{\max}_{\mathrm{global,capped}},
\]
where $c^{\max}_{\mathrm{global,capped}}$ denotes the capped global maximum attention score used to reduce the influence of extreme outliers. 
In WebShop, we set
\[
\theta_{\mathrm{low}} = 0.9 \cdot c^{\max}_{\mathrm{global}} .
\]
Segments whose utilization scores fall below the corresponding threshold are treated as weakly utilized, and are simplified only when they are also identified as semantically redundant by the Semantic Reflector.

For re-execution verification in the failure branch, we set the fixed maximum number of refinement attempts to 3 for ALFWorld and SciWorld, and 2 for WebShop. 
If no candidate update passes re-execution within this limit, AGMR rolls back to the original memory entry.

For verification purposes, an update is also accepted if the re-executed trajectory yields a reward 0.3 higher than that of the original trajectory. For WebShop, this threshold is set to 0.0.

\subsection{Other implementation details}
\label{app:other_implementation_details}

For WebShop, observations are usually much longer than those in ALFWorld and SciWorld. 
To avoid retaining verbose webpage content in refined memory, AGMR masks observation fields during simplification. 
This implementation choice partly explains why AGMR substantially reduces the average memory-token count relative to Traj on WebShop.

The initial trajectory-memory base used for AGMR offline refinement is extracted from the SFT dataset released by ETO~\cite{song2024trial}. 
All memory entries are drawn from training tasks, and the refined memory base is then evaluated directly on the test set.


\section{Prompt templates}
\label{app:prompt_details}

\begin{tcolorbox}[
  breakable,
  colback=gray!5,
  colframe=black,
  boxrule=1pt,
  arc=6pt,
  left=8pt,
  right=8pt,
  top=6pt,
  bottom=6pt
]
\textbf{Failure-branch Semantic Reflector prompt template, illustrated with SciWorld tasks.}

\begin{Verbatim}[fontsize=\footnotesize,breaklines=true]

[Agent Action Trajectory]
omitted
[Successful Memory Trajectory (Expert Trajectory)]
omitted
*Task Description*

You are given two trajectories for the same ScienceWorld laboratory task:

- Agent Action Trajectory: a step-by-step sequence of Thought–Action pairs produced by an agent, which ultimately fails to complete the task.
- Successful Memory Trajectory (Expert Trajectory): a reference trajectory in which the task is successfully completed.

Your task is to perform a structured failure analysis by aligning the Agent Action Trajectory with the Expert Trajectory.

The objective is:

1. Construct an Optimal Trajectory based ONLY on the Expert Trajectory.
2. Identify the FIRST erroneous step in the Agent Action Trajectory via stepwise alignment with the Optimal Trajectory.
3. Provide the Corrected Step that the agent should have taken, strictly grounded in the Expert Trajectory.
4. Identify the step in the Expert Trajectory that corresponds to the Corrected Step, and output it as the Expert Trajectory Reference.

------------------------------------------------------------

Section 1 (MANDATORY): OPTIMAL TRAJECTORY CONSTRUCTION

Before performing any failure analysis, you MUST first construct an Optimal Trajectory.

Definition: 
The Optimal Trajectory is the minimal canonical execution sequence that:

- successfully completes the task
- contains no mistakes or redundant steps
- follows the true logical task order (not environment-specific details)

*Core Rule (CRITICAL)*
The Optimal Trajectory MUST be derived exclusively from the Successful Memory Trajectory (Expert Trajectory).

This means:
- The Optimal Trajectory should follow the exact action flow of the Successful Memory Trajectory (Expert Trajectory).
- The Optimal Trajectory should closely mirror the sequence of actions in the Expert Trajectory.
- The only permitted modification when constructing the Optimal Trajectory is removing task-specific or environment-specific details
- The Optimal Trajectory should be viewed as a cleaned and generalized version of the Expert Trajectory, not a newly reasoned solution.
- You MUST NOT use the Agent Action Trajectory as a reference for constructing the Optimal Trajectory. Because the Agent Action Trajectory represents a failed attempt and may contain incorrect reasoning or actions.

Output Requirement

The Optimal Trajectory represents the perfect reference execution.
All subsequent failure analysis MUST be performed exclusively relative to this trajectory.

Example:
- Step 1: Teleport to the workshop and locate the unknown substance T.
- Step 2: Pick up the unknown substance T and focus on it.
- Step 3: Connect the battery anode to the first wire (e.g., green wire terminal 1).
- Step 4: Connect the battery cathode to the second wire (e.g., red wire terminal 1).
- Step 5: Connect the other end of the first wire (green wire terminal 2) to a terminal of a light bulb (e.g., cathode).
- Step 6: Connect the other terminal of the light bulb (anode) to a third wire (e.g., orange wire terminal 2).
- Step 7: Connect terminal 1 of the unknown substance T to the second wire (e.g., red wire terminal 2).
- Step 8: Connect the other end of the third wire (green wire terminal 1) to terminal 2 of the unknown substance T.
- Step 9: Wait and observe if the light bulb turns on.
- Step 10: Move the unknown substance T to the green box if it is conductive (bulb on), or to the blue box if it is nonconductive (bulb off).

------------------------------------------------------------

Section 2: Erroneous Step Localization Objective

You must identify the FIRST erroneous step via sequential alignment.

The Erroneous Step is defined as:
- the earliest point where the agent’s thought or action deviates from the Optimal Trajectory.
Only the first deviation may be selected.

Erroneous Step localization must be performed through sequential alignment.

Procedure
- Start from Step 1 of the Optimal Trajectory.
- Check whether the agent correctly completes this step.
- If correct, move to the next step.
- Stop when the first mismatch between the agent behavior and the required next step appears.

This step is the Erroneous Step.

Key Principle
After completing Step i, the agent should attempt Step i+1.
Any thought, action, or decision that does not aim to complete Step i+1 is considered a deviation.

A step is also an Erroneous Step if the agent makes a critical mistake that prevents task success, including:
- a key misjudgment that blocks a viable path to completion
- violating a required output format so the environment cannot interpret the response

If, after performing the stepwise alignment procedure, you cannot determine the exact step at which the first deviation occurs, or you have no clear idea about the failure reason, you MUST skip the full failure analysis and output exactly:

[Object not found]

This is the only allowed alternative to the STRICT OUTPUT FORMAT below.

------------------------------------------------------------

Section 3: Corrected Step And Expert Trajectory Reference

The Corrected Step represents the exact action the agent SHOULD have taken at the failure point to remain consistent with the Optimal Trajectory.

The Expert Trajectory Reference Thought–Action pair MUST be the step in the Expert Trajectory that corresponds to, or most directly informs, the Corrected Step.

------------------------------------------------------------

STRICT OUTPUT FORMAT (NO DEVIATION)

Your output MUST strictly follow the format below.
Do NOT add, remove, or reorder sections.
If the Special Case applies, output ONLY: Object not found.

[Optimal Trajectory]
- Step 1: <description>
- Step 2: <description>
- ...

[Step-by-Step Failure Reasoning]
<Sequential comparison between the agent trajectory and the Optimal Trajectory,
ending exactly at the first detected deviation>

[Erroneous Step]
Thought: <original agent thought>
Action: <original agent action>

Failure Reason:
<Why this step violates the Optimal Trajectory and what should have happened instead>

[Corrected Step]
Correct Thought: <thought aligned with the Optimal Trajectory>
Correct Action: <action aligned with the Optimal Trajectory>

[Expert Trajectory Reference]
Referenced Thought: <expert thought>
Referenced Action: <expert action>

\end{Verbatim}
\end{tcolorbox}

\begin{tcolorbox}[
  breakable,
  colback=gray!5,
  colframe=black,
  boxrule=1pt,
  arc=6pt,
  left=8pt,
  right=8pt,
  top=6pt,
  bottom=6pt
]
\textbf{Failure-branch Attention-Guided Refiner prompt template, illustrated with SciWorld tasks.}

\begin{Verbatim}[fontsize=\footnotesize,breaklines=true]
You are given the following inputs from a failed ScienceWorld task execution:

- [Erroneous Step Context]: a trajectory excerpt containing the failure point

- [LLM Analysis Response], which includes:
  - [Erroneous Step]: the critical step that caused task failure
  - Failure Reason
  - [Corrected Step]: the step that should have been executed instead
  - [Expert Trajectory Reference] (segment[8]): the expert segment that defines the correct behavior

- [Attention Score Distribution]: the LLM’s attention weights over Expert Trajectory segments at the moment the Erroneous Step was produced

The task is identical or highly similar to the expert trajectory. The failure point and its correct reference have already been identified.

Your task is to:

1. Analyze the attention distribution to determine which segments the LLM actually relied on when producing the erroneous Thought and Action.
2. Explain how this attention allocation led to the incorrect reasoning and behavior based on LLM Analysis Response and Erroneous Step Context.
3. Modify the relevant segment to prevent similar failures in future executions.

[Erroneous Step Context]
From [Task Trajectory]. Context window includes up to 3 Thought/Action/Observation groups immediately before the Erroneous Step.
Context window: 3 step(s) before the Erroneous Step.
omitted
    
=== LLM Analysis Response ===
omitted

==================================================

[Attention Score Distribution]
At the erroneous step (step 8), the LLM's attention distribution across memory segments was:
omitted

==================================================

[Attention Analysis Guidelines]

Calculate the average attention score: 63.0237
Expert segment[8] received attention score: 242.4957
Max attention score at erroneous step: 242.4957 (at segment[8])
Global maximum attention score in full matrix: 407.8840
High attention definition: score > 1.5× average AND score > 0.5× global maximum
High attention thresholds: > 94.5356 (1.5× average) and > 203.9420 (0.5× global maximum)

Based on the attention distribution, determine which scenario applies:

**Scenario 1: High attention to expert segment, but still failed**
- If segment[8] has high attention ( > 1.5× average AND > 0.5× global maximum )
- AND no other segment has comparably high attention
- This suggests the LLM did reference the expert segment but it may contain errors or insufficient emphasis
- Action: Modify segment[8] based on the Analysis Response to correct any errors and strengthen the guidance

**Scenario 2: Low attention to expert segment, no strong attention elsewhere**
- If segment[8] does NOT satisfy the high-attention definition
- AND no other segment meets the high attention definition (all scores relatively uniform)
- This suggests the LLM failed to retrieve or recognize the expert guidance, and no strong competing signal existed.
- Action: Modify segment[8] to make it more prominent and easier to retrieve (add emphasis, keywords, clearer structure)

**Scenario 3: Other segments have high attention (attention competition / distraction)**
- Regardless of whether segment[8] itself has high or low attention,
- If one or more segments other than segment[8] meet the high-attention definition ( > 1.5× average AND > 0.5× global maximum ), especially when those high-attention non-expert segments involve the same action type as the one taken in the [Erroneous Step] (e.g., both involve a take action)

- This suggests the LLM was misled by irrelevant segment[s] that competed for attention
- High-attention segments (excluding expert segment): segment[9] (208.4000)
- Action: Modify BOTH the expert segment AND the misleading segmen[s]:
  - Strengthen segment[8] to make it more relevant and prominent
  - Modify the misleading segment[s] to reduce irrelevant information or clarify their scope

[Important Notes]

- The attention score distribution may vary in complex ways beyond these three scenarios
- Use your judgment to analyze the actual attention pattern
- Focus on understanding WHY the LLM failed at the mechanism level
- Modify segments to address the root cause identified through attention analysis

==================================================

[OUTPUT FORMAT — STRICT, NO DEVIATION]

Step 1: First, provide your step-by-step analysis

[Step-by-Step Analysis]
1. Attention Pattern Analysis:
   <Analyze which scenario applies based on the attention scores>

2. Root Cause Identification:
   <Identify why the LLM made the error at the mechanism level, and determine WHICH segment(s) require modification as a direct result of this error>

3. Modification Strategy:
   <Explain HOW the identified segment(s) will be modified.
   The strategy should primarily focus on the erroneous *Thought* in the [Erroneous Step].
   It should describe how the incorrect reasoning or intention expressed
   in that Thought can be corrected, constrained, or redirected,
   with the explanation grounded in both the attention pattern analysis
   and the error analysis provided in the === LLM Analysis Response ===.>

Step 2: Then, output the modified segments

For EACH segment you modify, output the modified content in the following format:

enhanced_segment[SEGMENT_ID]: <modified segment content>

Before the "[/INST] Thought" section in enhanced_segment, you must output a text segment in the following format:
"*IMPORTANT NOTE*: I MUST <do the correct action> In this environment... If the agent performs <wrong action>, the episode immediately fails."


Example:
enhanced_segment[3]: [INST] Observation: This outside location is called the outside. Here you see: 
	the agent
	a substance called air
	a blue jay egg
	a giant tortoise egg
	a substance called wood
You also see:
	A door to the greenhouse (that is open)
	A door to the kitchen (that is open) *IMPORTANT NOTE*: I MUST enter the menu index (e.g., Action: 0) if "Ambiguous request: Please enter the number for the action you intended" appears. In this environment, if a numbered list appears, repeating the text command causes an error. If the agent performs a text repetition, the episode immediately fails. [/INST] Thought: I see several eggs outside, which are animals. I will focus on the blue jay egg. If the ambiguous prompt appears, the Action must output 0 instead of repeating the command.
Action: focus on blue jay egg

--------------------------------------------------
Formatting rules:
- Do NOT use quotation marks
- Preserve the Observation/Thought/Action structure from the original segment
- Do NOT output anything else besides the [Step-by-Step Analysis] section and enhanced_segment lines
- When writing enhanced_segment content, prefer simple, direct words and sentences.
- Use concise, command-style wording whenever possible to clearly specify the required action.
- Avoid long or complex explanations inside enhanced_segment.
==================================================
Length Constraint for each enhanced_segment:
- Apply minimal edits.
- Do not significantly increase the segment length.
- Keep enhanced_segment within 150% of the original segment length. 
- Avoid redundant explanations.

\end{Verbatim}
\end{tcolorbox}

\section{Computation resources}
\label{app:computation_resources}
All experiments are conducted on NVIDIA RTX PRO 6000 GPUs with 97887MiB memory. The peak GPU memory consumption varies with the model size. Specifically, Qwen-7B and Llama-8B generally require less than 24 GB of VRAM. The runtime of AGMR ranges from 4 to 16 hours, which depends on the backbone model, benchmark task, and API response speed.


\end{document}